# Explainable Artificial Intelligence in Construction: The Content, Context, Process, Outcome Evaluation Framework

Peter E.D. Love[a], Jane Matthews[b], Weili Fang[c,d], Stuart Porter[e], Hanbin Luo[f], and Lieyun Ding[g]

[a] School of Civil and Mechanical Engineering, Curtin University, GPO Box U1987, Perth, Western Australia 6845, Australia, Email: p.love@curtin.edu.au

[b] School of Architecture and Built Environment, Deakin University Geelong Waterfront Campus, Geelong, VIC 3220, Australia, Email: jane.matthews@deakin.edu.au

[c] Department of Civil and Building Systems, Technical University of Berlin, Gustav-Meyer-Allee 25, 13156 Berlin, Germany, Email: weili.fang@campus.tu-berlin.de

[d] School of Civil and Mechanical Engineering, Curtin University, GPO Box U1987, Perth, Western Australia 6845, Australia, Email: weili.fang@curtin.edu.au

[e] School of Civil and Mechanical Engineering, Curtin University, GPO Box U1987, Perth, Western Australia 6845, Australia, Email: stuart.porter@curtin.edu.au

School of Civil Engineering and Mechanics, Huazhong University of Science and Technology, Wuhan, 430074, China, Email: Luohbcem@hust.edu.cn

School of Civil Engineering and Mechanics, Huazhong University of Science and Technology, Wuhan, 430074, China, Email: dly@hust.edu.cn

# Explainable Artificial Intelligence in Construction: The Content, Context, Process, Outcome Evaluation Framework

**Abstract**

Explainable artificial intelligence (XAI) is an emerging and evolving concept. Its impact on construction, though yet to be realized, will be profound in the foreseeable future. Still, XAI has received limited attention in construction. As a result, no evaluation frameworks have been propagated to enable construction organizations to understand the what, why, how, and when of XAI. Our paper aims to fill this void by developing a *content*, *context*, *process,* and *outcome* evaluation framework that can be used to justify the adoption and effective management of XAI. After introducing and describing this novel XAI framework, we discuss its implications for future research. While our novel framework is conceptual, it provides a frame of reference for construction organizations to make headway toward realizing XAI's business value and benefits.

## 1.0 Introduction

Artificial intelligence (AI) technologies are evolving rapidly and profoundly impact the decision-making and operations of a business. The exponential growth in AI and widespread diffusion of applications into the workplace has left many people questioning and seeking to understand *why* and *how* recommendations or decisions are made. Moreover, a debate about bias and fairness, and trust and reliability has arisen with AI, especially as we see it increasingly used in areas such as recruitment, real-time progress monitoring, cyber-security, risk management, and safety (Arrieta *et al*., 2020; Angelov *et al*., 2021; Love *et al*., 2022). However, human decision-making in these domains, amongst many others, can also be flawed and prone to errors or bias, as an extensive body of research has demonstrated (Kahneman *et al*., 1982; Hilbert, 2012).

While traditional Machine Learning (ML) algorithms, also known as *transparent models* such as linear regression, decision trees, and Bayesian networks, are intuitive and explainable, their accuracy can be questioned. Consequently, we have seen *opaque models* such as Support Vector Machines (SVM), Random Forest, and Deep Learning (DL) gain prominence to counteract the issues of accuracy, but this has come at the expense of explainability (Arrieta *et al*., 2020; Angelov *et al*., 2021; Sokol and Flach, 2022). The loss of explainability in AI models has resulted in Gerling *et al*. (2021) pointing out that "if we cannot explain the algorithms, we cannot argue against them, verify them, improve them, or learn from them". So, if we can generate explanations to help understand how DL and ML models function and why decisions or predictions are made, trust in them can be established. In a nutshell, XAI is a set of processes and methods that allow us to understand and trust the results and output of DL and ML algorithms.

Within the construction literature, limited attention has been given to XAI (Naser, 2021; Geetha and Sim, 2022; Love *et al*., 2022). Construction organizations remain at the cusp of the AI revolution, which has now taken off in other industrial sectors such as healthcare, manufacturing, and transport. The hesitancy in adopting AI is not so much to do with being reluctant but instead lacking an understanding and knowing *what*, *why,* and *how* its benefits can be realized without compromising its operations (Love *et al*., 2022). Unquestionably, AI-enabled technologies such as autonomous plant, robotics, unmanned aerial vehicles (UAV), three-dimensional (3D) printing, and the Internet of Things (IoT) will provide construction organizations and projects with significant productivity and performance improvements. We submit that the future looks bright for the adoption of AI if we can demonstrate the benefits of XAI to construction organizations.

With the need for explainability becoming a critical feature for the deployment of AI (Arrieta *et al*., 2020), construction organizations are well-positioned to embrace this perspective as they have limited AI baggage, only relying on a few proprietary applications (e.g., Autodesk BIM 360 Workflow), if any, to help them perform and manage their operations. An essential goal of XAI is algorithmic accountability, as many AI systems have traditionally been black boxes. Even when the inputs and outputs are known, the algorithms used to make decisions are often proprietary and not easily understood (Anand *et al*., 2020).

While XAI precepts, methods (including post-hoc explainability), and techniques are evolving as a consequence of an intense period of exploration and debate in the literature, limited attention has been given to its evaluation (Langer *et al*., 2021; Rosenfeld, 2021; Love *et al*., 2022; Meske *et al*., 2022). In filling this void, this paper aims to develop a holistic theoretical framework that can be used as a frame of reference to evaluate XAI and advance its adoption in construction. At this juncture, we would like to point out that technology evaluation has received limited attention in construction literature (Love and Matthews, 2019), despite its critical role in ensuring its successful adoption across its life cycle (Seddon, 1997; Irani, 2002; Irani *et al*., 2003; Meske *et al*., 2022).

As AI models become more complex and interconnected, their *evaluation* becomes even more necessary as we seek to understand how well they achieve their goals and determine their effectiveness and explainability. Technology evaluation is not the mainstay of a construction organization's business strategy, as its immediate goal is to deliver the physical end product within pre-defined deliverables. However, construction organizations are now beginning to realize the importance of managing and utilizing their data for decision-making (i.e., *what*, *how,* and *when*), as it can result in successful project outcomes and provide them with a competitive advantage (Matthews *et al*., 2022).

In response to the calls to develop XAI evaluation frameworks due to their absence in the literature (Langer *et al*., 2021; Rosenfeld, 2021; Love *et al*., 2022; Meske *et al*., 2022), this paper aims to address this research gaps. In doing so, we propagate a new evaluation framework for XAI. In this instance, the framework will address what is being evaluated, why the evaluation is being undertaken, those factors that influence the evaluation, how it will be carried out, and its explainability.

We commence our paper by examining the nature of technology evaluation and its importance in ensuring its benefits are realized (Section 2). We then introduce and describe the Context, Content, and Process (CCP) framework developed initially by Pettigrew (1985) to study organizational change and then reframed to accommodate technology evaluation by Stockdale and Standing (2006) and Stockdale *et al*. (2006) (Section 3). Building on the earlier technology evaluation research and taking into account the nuances and context of XAI, a new and bespoke evaluation framework is proposed (Section 4). The implications for research going forward are then discussed (Section 5) before concluding our paper (Section 6).

## 2.0 Technology Evaluation

Evaluation is the process of gathering information designed to assist decision-making about the object to be evaluated (Owen, 2006). Several definitions of technology evaluation can be found in the literature focusing on varying contexts (Irani and Love, 2008). For example, Remenyi *et al*. (1997) focus on the determination of the worth and value of technology by judging it in accordance with specific criteria and, in doing so, offer the following definition:

> "A series of activities incorporating understanding, measurement, and assessment. It is either a conscious or tacit process that aims to establish the value of or the contribution made by a particular situation. It can also relate to the determination of the worth of an object" (p. 46).

Contrastingly, Farbey *et al*. (1993) offer a different slant defining technology evaluation as a "process that places at different points in time or continuously, for searching and for and making explicit, quantitatively or qualitatively, all of its impacts (p.205). After decades of research, there is widespread consensus that the focus of evaluation should not be purely on technology but also on its complex interactions with an organization's social and political dimensions and stakeholders (Hirschheim and Smithson, 1988; Symons, 1991; Farbey *et al*., 1993; Walsham, 1993; Smithson and Hirschheim, 1998; Irani and Love, 2000; Irani and Love, 2002; Love *et al*., 2004; Stockdale *et al*., 2006; Andargoli *et al*., 2017; Pereira *et al*., 2022).

Markedly, Hirschheim and Smithson (1988) point out that if an evaluation focuses solely on technology dimensions, then meaningless conclusions will materialize. So, as XAI is a *meta-human system* (i.e., humans + machines that learn) – a form of socio-technical system – we need to consider the relations between people, technology, tasks, and the organization when developing an evaluation framework for its use in practice (Lyytinen e*t al*., 2020; Gerlings *et* *al.*, 2021). Thus, when creating an evaluation framework for XAI, we need to recognize the importance of stakeholders and consider their desiderata (i.e., requirements) (Amarasinghe *et al.*, 2021; Langer *et al.*, 2021; Love *et al.*, 2022). The structure for XAI in construction (XAI-CO) with critical questions to be considered during an evaluation process is presented in Figure 1.

We hasten to note that the evaluation of technology is a challenging and complicated process due to its ever-changing advancements (i.e., increased accuracy, scope, functionality, and flexibility due to developments in AI) and impact on organizational infrastructure (Irani, 2002; Andargoli *et al.*, 2017). Considering this pervasive challenge, the information systems literature is replete with studies that have sought to address technology evaluation over the past three decades, proposing various frameworks, methodologies, and metrics (Symons, 1991; Farbey *et al.*, 1993; Walsham, 1993; Smithson and Hirschheim, 1998; Irani and Love, 2000; Gunasekaran *et al.*, 2006; Andargoli *et al.*, 2017; Pereira *et al.*, 2022). Before considering a framework, methodology, and metrics for evaluation, a decision must be made to justify an investment in technology.

## *2.1 Investment Justification*

An investment in technology, such as an AI-enabled system, can be expensive for construction organizations and projects. So, like any investment decision, organizations typically need to undertake some form of justification and demonstrate that the expected benefits (e.g., strategic, operational, and tactical) realized from implementing AI can be achieved. The investment justification process is by no means straightforward, as the direct and indirect costs, risks, and benefits (including dis-benefits) all need to be considered (Love and Matthews, 2019).

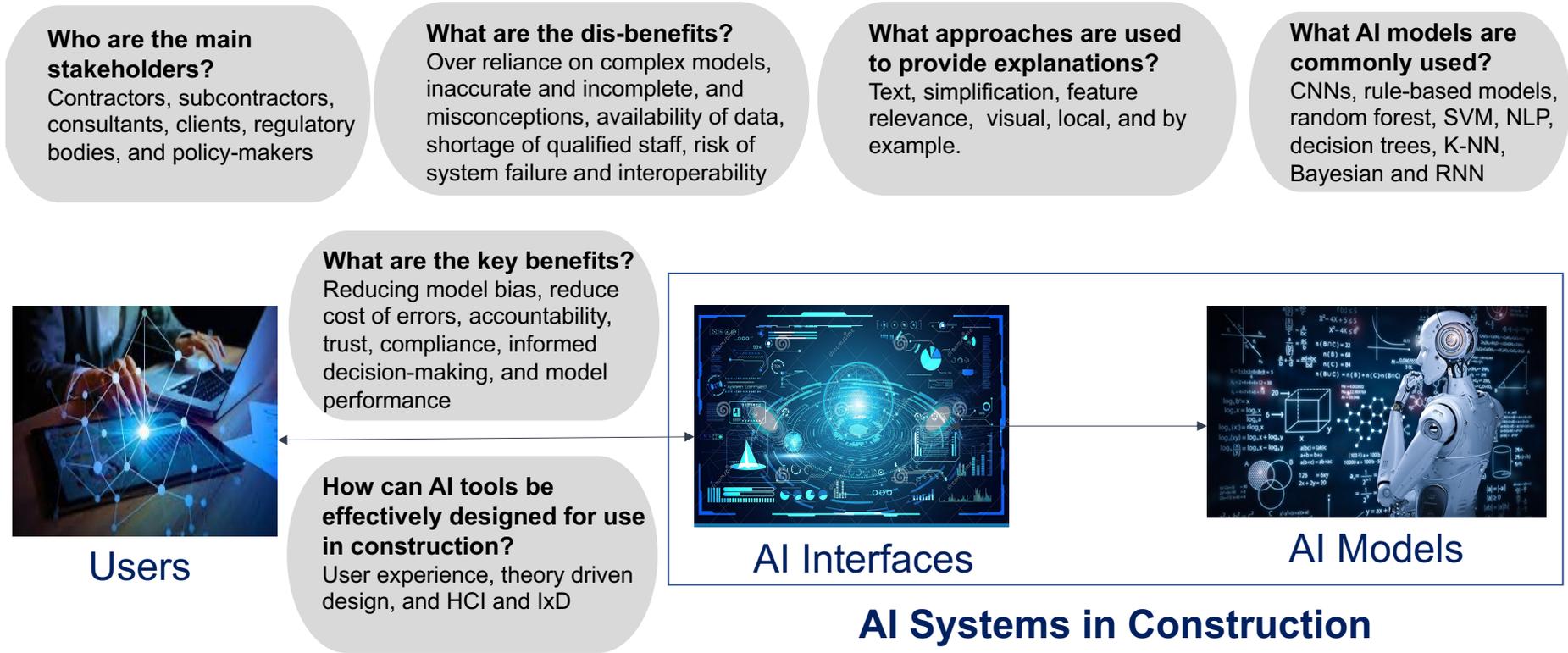

Adapted from Khosravi *et al.* (2022: p.4)

Figure 1. The structure for XAI in construction

Nevertheless, construction organizations tend to justify their investments in technology using traditional measures such as the Return on Investment (RoI) or even as an act of faith to keep up with and mimic their competitors (Love and Irani, 2004; Barlish and Sullivan, 2012; Love *et al*., 2013). If measures such as "RoI are used for justifying the adoption of technology, then the process evaluation only serves the objective of management" as it becomes a financial exercise (Love *et al*., 2013: p.209).

Should a decision to implement AI-enabled systems be undertaken based on RoI, then stakeholder desiderata would be ignored. The upshot being they may be unwilling to collaborate in realizing the potential of the AI system. While AI solutions can bolster performance and productivity improvements and increase profits, they can also add value by providing organizations with the 'business intelligence' needed to sustain a competitive advantage. With a shift toward the use of collaborative procurement forms (e.g., Alliancing and Integrated Project Delivery) that aim to integrate their supply chains, there is a growing need to engender 'business intelligence' in projects to enable access to and analysis of information (including stakeholders) to improve and optimize decisions and maximize an asset's performance over its life (Love *et al*., 2021; Matthews *et al*., 2022).

In the foreseeable future, we may no longer view construction organizations as managers of manually-laden, labor-intensive processes used to build unique structures and physical infrastructure. However, instead, we could see 'intelligent constructors' utilizing a suite of cyber-physical systems (e.g., 3D printing and UAV) to enable and manage smart, connected, integrated, and automated operations (e.g., progress monitoring, component tracking, and control, and crane safety) to deliver an 'intelligent asset'. Unquestionably AI is critical for the future of work in construction. However, for its benefits to be realized, explainability is a *sine qua non* (Love *et al*., 2022).

## 2.2 Benefits Realization

Benefits management comprises a range of activities designed to ensure that a construction organization and its projects realize the benefits it plans to achieve from implementing AI. The construction literature is replete with reviews identifying the benefits and opportunities for AI research (Darko *et al*., 2020; Debrah *et al*., 2022; Pan and Zhang, 2021; Zhang *et al*., 2022). Despite its importance, missing from these reviews is a reference to XAI. Thus, we question the legitimacy of the presented benefits and research opportunities as explainability is ignored.

We have yet to understand and realize the business benefits of XAI in construction as it has received limited attention. But like any new technology, benefits cannot be delivered without change, and change without benefits cannot be sustained (Love and Matthews, 2019). In the case of construction, this change centers on organizations preparing themselves for a new way of working (e.g., education, upskilling, and new processes) where cyber-physical systems are viewed as key enablers to improved and sustained business performance. A detailed description of the 'how' of benefits management for digital technology can be found in Love and Matthews (2019).

Indeed, the benefits (i.e., strategic, operational, and tactical) of AI can be potentially exponential for construction organizations and include: (1) time and cost savings (also productivity dividends) by automating and optimizing routine processes and tasks; (2) quicker businesses decisions based on outputs from cognitive technologies (e.g., computer vision, robotics and voice recognition); (3) mining large amounts of data that can be used to generate data analytics; (4) improved identification of safety and quality risks; and (5) increased profitability.

However, as we have mentioned, an AI model's ability to deliver benefits will be limited in construction unless the precepts of XAI are addressed (Arrieta *et al*., 2020; Angelov *et al*., 2021; Love *et al*., 2022): (1) *explainability*, which describes the active property of a learning model based on *posteriori* knowledge and aims to clarify its functioning; (2) *interpretability*, which refers to the passive property of a learning model based on a *priori* knowledge to ensure that a given model makes sense to a human observer using language that is meaningful to the user; and (3) transparency (including simulatability, decomposability, and algorithmic transparency). Post-hoc explanations for outputs may also be needed, which can take various forms such as text explanations, visual explanations, local explanations, explanations by example, explanations by simplification, and feature relevance explanations (Figure 1). A detailed description of post-hoc explainability methods can be found in Lipton (2018), Arrieta *et al*. (2020), Belle and Papantonis (2021), and Speith (2022).

Determining the appropriate AI algorithm[1], explanation technique, and method for presenting the explanation are outside the scope of this paper. However, we need to point out that they can influence the effectiveness of the benefits realization process. As XAI makes it possible to open up the algorithmic 'black box' of DL and ML models, meaningful explanations, additional applications, and software considerations will undoubtedly be needed to support prevailing AI systems. Alternatively, explainability can be considered up-front and embedded into the design of a system, which will influence the algorithm selected and how it pre-processes data (Oxborough *et al*., 2018).

---

[1] Algorithms can catogerized as transparent (e.g., decision trees, K-NN, rule-based learners, Bayesian) and opaque (e.g., CNNs, RNN, random forest, and SVM) (Arrieta *et al*., 2020).

### 2.2.1 Building in Explainability: A Benefit in the Making, but at a Cost

Rather than adopting an unexplainable propriety AI system, construction organizations may consider designing and building their own bespoke interpretable, and inclusive systems from the ground up using approaches such as Human-Computer Interaction (HCI) or Interaction Design (IxD) (Figure 1) (Xu, 2019). Consequently, end-user trust and improved transparency with human-interpretable explanations can be ensured. Developing an AI system can be expensive, requiring considerable research and development, which will invariably be outside the remit of most construction organizations. However, if construction organizations develop their own AI systems designed to be explainable, they could be deployed confidently.

The business benefits of building explainability into AI systems include: (1) reducing the cost of mistakes; (2) reducing model bias; (3) a source of accountability for mistakes; (4) trust and reliability; and (5) understanding the model's performance. The greater the confidence in AI, the faster and more widely it can be deployed. Thus, when explainability is used with AI, it becomes a "competitive differentiator" (Oxborough *et al*., 2018: p.17).

### 2.2.1 Management of Dis-Benefits and Drawbacks

While construction organizations such as Obayashi (e.g., the development of AiCorb to rapidly design façades and use of DL to evaluate mountain tunnel faces) and Skanska (e.g., movement of heavy equipment and curbing equipment emissions) look for ways that AI can improve their productivity, performance, and business results, we need to acknowledge there are also disbenefits associated with its implementation.

Some of the dis-benefits of AI include: (1) availability of data (e.g., its volume, format, and quality); (2) shortage of qualified staff who have the skills and experience to deploy and operate

AI; (3) direct and indirect costs across the applications life including the integration with other technologies; (4) risks of system failure and loss of important data; and (5) usability and interoperability with other systems and platforms. Notwithstanding, there are ethical concerns that construction organizations may need to address, such as job losses due to automation, the need to redeploy or re-train staff, and security and privacy issues. Acknowledging the dis-benefits of AI will also help understand those associated with XAI, as they are not mutually exclusive.

More specifically, however, a mistake that people often make is to rely on complex ML models in cases where an interpretable model would have delivered a comparable or superior performance (Khosravi *et al*., 2022). For example, logistic regression has outperformed DL models on datasets of moderate size (Gervet *et al*., 2020). Thus, simple interpretable models should be initially used when designing an AI system. Then those of a more complex nature should be introduced in a "step-wise manner where performance in accuracy and interpretability are measured and compared (García-Rodríguez *et al*., 2022; Khosravi *et al*., 2022: p. 8). It is generally accepted that DL models provide better accuracy than their ML counterparts, with explainable deep neural networks (xDNNs) being the best performers (Angelov and Soares, 2020; Angelov *et al*., 2021).

Explanations that are misleading or inaccurate can result in unfavorable outcomes for stakeholders. Inaccurate explanations often materialize due to poorly developed models prone to under or over-fitting or noisy data. As an interpretation can only be as good as its underlying model, it must be rigorously developed. Its selection needs to be based on best practices. A process of hyperparameter tuning, followed by evaluation, will need to be undertaken before an AI system is adopted by a construction organization (Molnar *et al*., 2020; Khosravi *et al*., 2022; Love *et al*., 2022).

Incomplete or simplified explanations may sometimes manifest as the complexity of an AI system is disguised by developers. At face value, the simplicity of an explanation, even though incomplete, may be appealing to users, but this can result in a "false sense of comprehension" that can result in adverse outcomes (Khosravi *et al*., 2022: p.8). Similarly, users can misinterpret an outcome as they do not understand the explanations. Adopting Kulesza *et al*.'s (2015) principles of 'explanatory debugging'[2] –based on a two-way exchange of explanations between an end-user and ML system –could help end-users build useful mental models of a learning system's behavior and understand its functioning, thus complementing XAI.

Understanding an explanation's *context* is necessary for crafting compelling human-centric explanations (Srinivasan and Chander, 2020). However, rather than solely considering end-user views, explanations need to accommodate those of stakeholders if the context is to have a meaning within the framing of XAI (Arrieta *et al*., 2020; Langer *et al*., 2021; Khosravi *et al*., 2022; Love *et al*., 2022). Thus, the context in this instance considers who is providing the explanation, who is seeking the explanation, what its purpose is, and what is the use case (Srinivasan and Chander, 2020). Additionally, there is a need to understand the *content* of the explanation and what constitutes a reasonable explanation for a particular type of user and use case.

---

[2] Explanatory debugging is an approach whereby "the system explains to users how it made each of its predictions and the user then explains any necessary corrections back to the system Kulesza *et al*.'s (2015: p.126). The principles of 'explanatory debugging' focus two themes: (1) *explainability* - be sound, complete and don't overwhelm; and (2) *correctability* – be actionable, be reversible, always honor user feedback and incremental change matters.

Explanations of AI can facilitate conversation between stakeholders enabling a *process* of co-design and value co-creation, empowering them to make conscious decisions about the system's use and when to evaluate its performance (Paez, 2019; Khosravi *et al*., 2022). So, when considering the evaluation of XAI, we need to understand the varying socio-political contexts of stakeholders, which "demands an interpretative approach to allow for the deepening of understanding and generation of motivation and commitment" (Stockdale and Standing, 2006: p.1090; Stockdale *et al*., 2016). This backdrop provides a segue for introducing and describing the CCP evaluation framework in the next section of this paper.

## 3.0    The Content, Context, and Process Framework

We present the CCP evaluation framework in Figure 2, which can be used to support the evaluation of technology across multi-level contexts and processes even though it has not been applied to any studies in construction. However, for the most part, it provides a starting point for undertaking an inquiry into justifying a technology investment beyond the narrow quantification of costs and benefits to the analysis of opportunities and potential constraints (Symons, 1991; Stockdale *et al*., 2006; Andargoli *et al*., 2017). Accordingly, Stockdale and Standing (2006) suggest that adopting the CCP framework provides pathways for identifying new evaluation approaches. Thus, the CCP framework is not a generic solution but instead provides the basis to address all relevant questions regarding the social, technical, cultural, and political factors that can influence an effective evaluation of technology. We will use the foundations of the CCP framework to develop a bespoke one for XAI-CO (Section 4).

### *3.1    Content*

The content within the framework refers to "the particular areas of transformation under examination" (Pettigrew, 1987: p. 9). Understanding *what* is being measured is core to any evaluation study, but this is not as straightforward as it appears, as stakeholders and organizational contexts need to be considered (Figure 2) (Symons, 1991; Gable *et al*., 2008). A major challenge in evaluating XAI, for example, arises from the general absence of technology usage in construction and the complexity of the organization-project dyad. Here technology needs to serve two masters with various stakeholder interests accommodated, some of whom will have conflicting goals that need to be met. Naturally, trade-offs will have to occur, but striking the right balance depends on what context is measured and how a suitable yardstick is selected to evaluate a process. As the content, context and process are interdependent, understanding their dynamic interactions is needed to derive a holistic view of the evaluation process (Figure 2).

## *3.2    Context*

Context is defined as the "formal and informal setting in which a situation occurs, including the political, legal, geographical, historical, socio-cultural, environmental, institutional or managerial circumstances and/or the continual unfolding interaction between the situation and setting" (Brown, 2010: p.7). Reinforcing this point Love and Ika (2021) note that a context enables us to capture interactions over time – it is "both existing and emerging, initial and unfolding, or static or fluids which has significant repercussions for their comprehension" (p.3). The importance of understanding the context and its role in engendering technology innovations in organizations has been echoed in numerous studies (Orlikowski and Baroudi, 1991; Chiasson and Davidson, 2004; Scheepers *et al*., 2006; Standing *et al*., 2017; Love and Matthews, 2019).

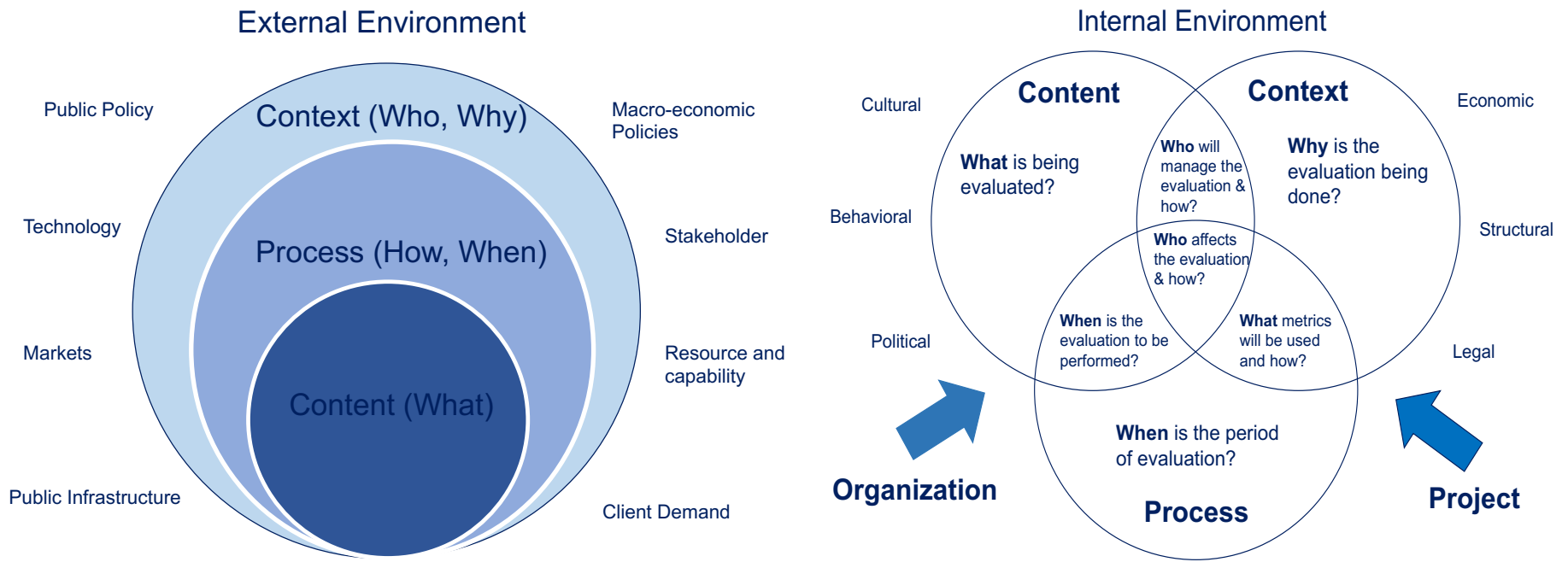

Adapted from Stockdale *et al.* (2006: p.733)

Figure 2. The content, context, and process model

As the context of the organization-project dyad is complex and very different from the standalone business context, it would be inappropriate to apply generic technology evaluation models to construction, no matter how persuasive they appear. Thus, in this instance, piggybacking off the CCP framework to develop a bespoke evaluation model for XAI provides a legitimate and plausible starting point for construction.

Understanding context has been critical to the success of implementing digital technologies in construction (Love and Matthews, 2019). However, when insufficient consideration is given to the evaluation of technology, then comprehending why it is successful or fails to deliver business benefits becomes a challenge. As we can see from Figure 2, the context comprises the external and internal environment, with each being impacted by factors that can influence the efficacy of an evaluation (Stockdale and Standing, 2006; Stockdale *et al*., 2006).

Two critical questions concerning context need to be addressed: (1) *why* is the evaluation being undertaken? and (2) *who* should be included in the process? Within the technology literature, 'the why of evaluation' has traditionally focused on the appraisal of value to measure success or recognize benefits (Irani and Love, 2002; Barlish and Sullivan, 2013; Love and Matthews, 2019). However, an evaluation can "reinforce an existing organizational structure for political or social reasons and be a ritualistic rather than effective process" (Stockdale and Standing, 2006: p.1093). The complexity of evaluation owes much to the different perceptions and beliefs of various stakeholders. Determining the stakeholders to be included in the evaluation of XAI will be complicated. However, we need to be aware that those holding power can skew the outcomes of the process to ensure their objectives are met (Jasperson *et al*., 2002; Stockdale *et al*., 2006; Andargoli *et al*., 2017).

## 3.3 Process

The process refers to interested parties' actions, reactions, and interactions as they attempt to change an organization from one state to another (Pettigrew, 1987). So, central to the evaluation process is an explanation of *how* and *when* it will be performed (Symons, 1991). Different lenses and approaches can be used to examine the 'how' of evaluation. From a philosophical perspective, the how may be evaluated through an objectivist or subjectivist lens (Walsham, 1993; Irani, 2002; Irani *et al.*, 2006; Andargoli *et al.*, 2017). Observations through an objective lens will reveal that different observers of the same phenomenon will arrive at the same conclusion. Objective methods and measures that derive quantitative outputs (e.g., simulation modeling, cost-benefit analysis, and RoI) will drive decision-making (Irani and Love, 2008). In contrast, those adopting a subjective and qualitative methods view will be unable to reach a consensus and derive a solution.

Since no research has focused on evaluating XAI in construction, a qualitative approach is most appropriate in this case to examine the process – answers to *why* and *how* can be provided – enabling deeper insights and greater assistance to decision-makers. Indeed, an objectivist evaluation approach is best suited for examining a system's performance (e.g., metrics). However, the unavailability of XAI models in use in construction puts this approach, for the time being at least, on the back burner. Nevertheless, if and when construction organizations adopt XAI, quantitative measures will need to be developed to evaluate a system's performance.

The *when* of evaluation depends on its purpose (Irani and Love, 2002). Evaluations are classified as (Farbey *et al.*, 1993; Irani and Love, 2002): (1) *ex-ante*, which focuses on justifying the investment decision; (2) *formative*, focusing on determining which aspects of the system's design works well or not, and why; (3) *summative*, focusing on how well a system performs, often compared to a series of benchmarks; and (4) *ex-post*, focusing on determining if benefits have been realized or not, and why. Symons (1991) argues for "treating [evaluation] as continuing throughout the various stages of system development" (p.211).

While the idea of continuously evaluating technology sits well within the academic community, it does not resonate with practitioners due to issues of cost and resourcing (Willcocks, 1992; Irani *et al*., 2003; Love *et al*., 2004; Berghout *et al*., 2011; Love and Matthews, 2019). However, by taking a life-cycle approach, we can consider how XAI influences and initiates changes to the organization and project context. This approach reduces the risk of failure by putting in place 'check points' to harness understanding and learning about a model's performance (Irani *et al*., 2003; Love and Matthews, 2019).

## 4.0     Extending the CCP: A Conceptual XAI Evaluation Framework

The concepts embedded in the CCP are used to develop a conceptual XAI evaluation framework, which we present in Figure 3. Noticeably, we have added the concept of 'outcome' (what) to the CCP framework so that the benefits of implementing XAI are explicit within the organizational and project context and to stakeholders. Our framework aims to capture the context and nuances of XAI, especially those of stakeholders (Table 1). For example, issues of explainability that a construction organization may consider enacting throughout the various phases of an XAI development and implementation are identified in Figure 4. With regards to content, important questions that need to be asked by a construction organization include: (1) 'what is the aim of introducing XAI and how can a project benefit from its use?'; (2) 'why is an explanation needed to understand a model?'; and (3) 'what function will explanations serve?'

We identify explainability approaches juxtaposed with their purpose that can be used to support the questions raised in Figure 4. For example, a surrogate model approach can be used to improve the interpretability of the DL (i.e., black box) model's output by comparing them to transparent ones[3]. To this end, we need to be able to explain (McNamara, 2022):

- What data went into a training model and why? How was fairness assessed? What effort was made to reduce bias? (*Explainable data*)
- What model features were activated and used to achieve a given output? (*Explainable predictions*)
- What individual layers make up the model, and how do they result in an output (*Explainable algorithms*)

Against the contextual backdrop presented above, we now provide a brief overview of how our framework could be operationalized.

## 4.1 XAI Framework Operationalization

At this point, we assume a construction organization has justified its investment in XAI and is wholeheartedly committed to incorporating it into its business functions and projects. As mentioned, the XAI evaluation framework in Figure 3 consists of the content, context, process, and outcome. Here the content is self-explanatory as we are dealing with XAI at the organization and project levels. However, it cannot be considered in isolation, as shown in Figure 1. The context is the underlying framing within which the construction organization (i.e., implementation setting) and the projects (i.e., situational setting) are delivered. Naturally, change and outcomes will emerge from the introduction of XAI.

---

[3] If a model is transparent, then a person understands the entire model, which implies it is uncomplicated. So, for a model to be understood, a person should be able to take the input data juxtaposed with the model's parameters and produce a prediction.

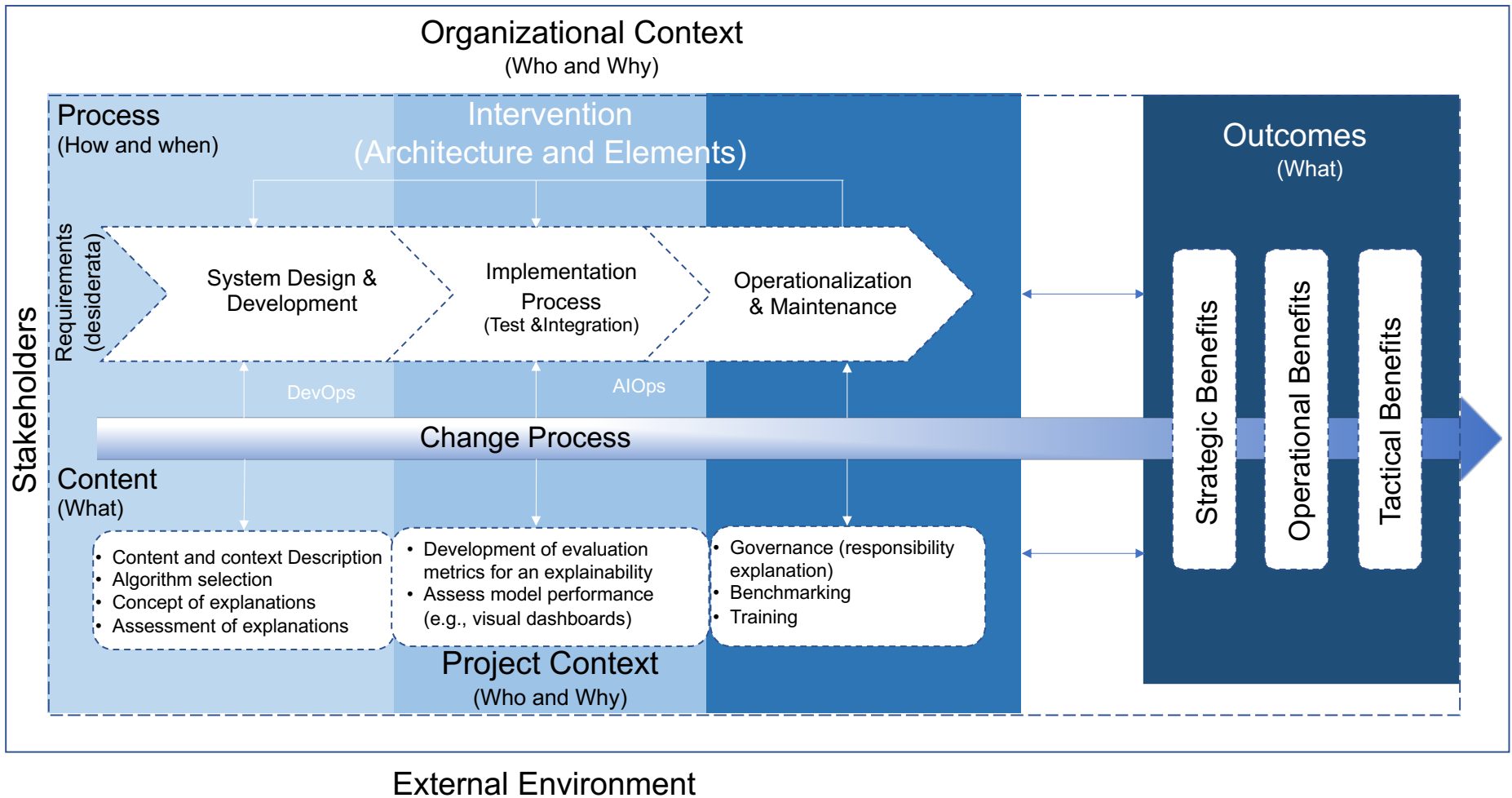

Adapted from Fridlich *et al*. (2015)

Figure 3. Conceptual XAI evaluation framework for construction organizations

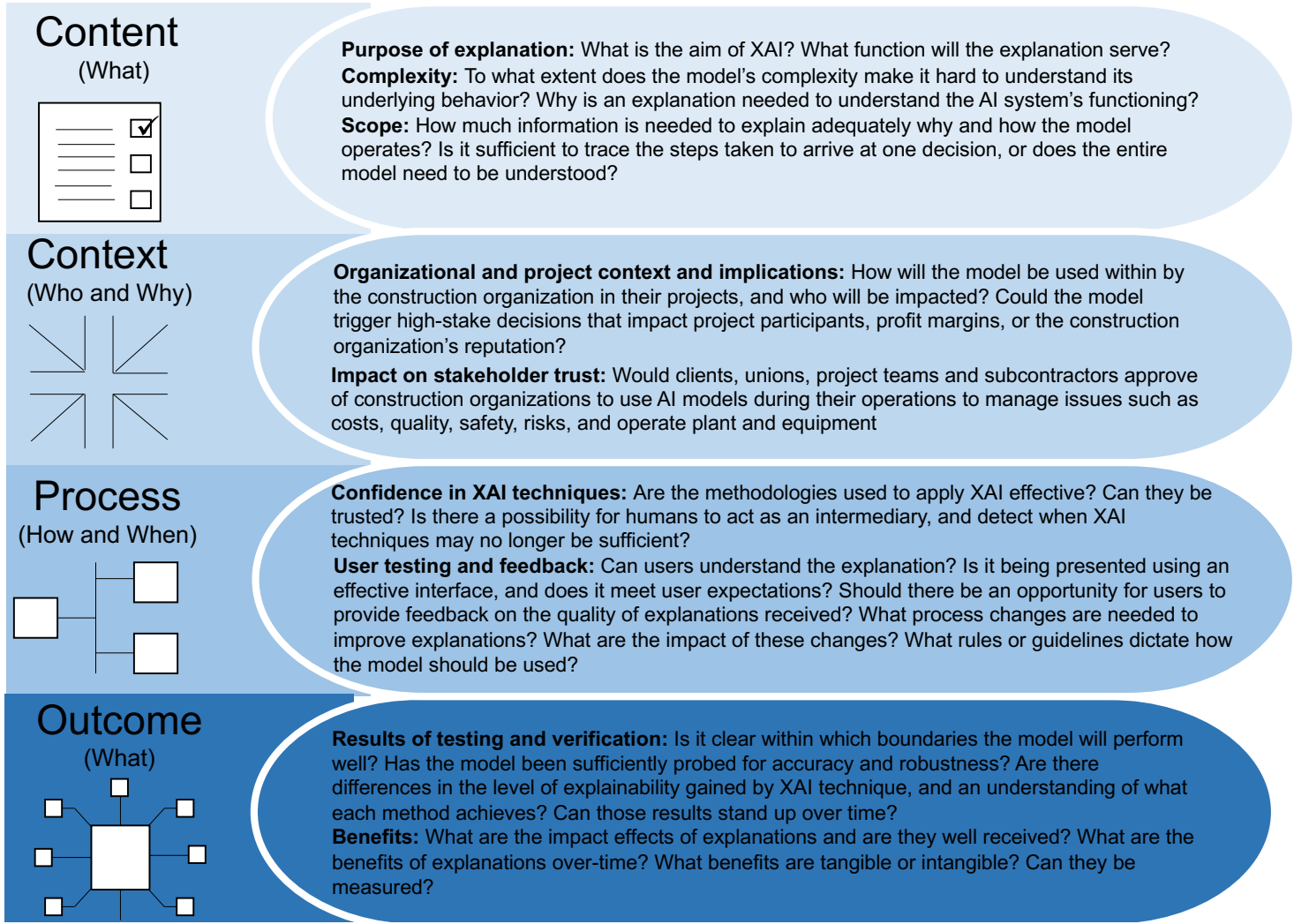

Figure 4. Examples of explainability issues a construction organization may consider enacting

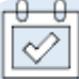

Figure 5. Different explainability approaches to be used as part of the XAI evaluation process

Table 1. A hypothetical example of a stakeholder lens to evaluate explainability methods in construction

| Stakeholder Role | Purpose | XAI Methods |
| --- | --- | --- |
| Software developer and data scientist | Weigh trade-offs between explainability and accuracy; assess how models perform—use real empirical data for training models. | Feature performance over time, which can reveal how the weighting of variables may change in new environments and conditions (e.g., project types) and under changing constraints |
| Executives, managers (e.g., construction, quality, risk), site supervisors, engineers | Test the robustness of models under varying conditions | Adversarial techniques that test whether tweaks in the input data have an undesirable impact on predictions |
| Design team (e.g., architect and engineers) and subcontractors | Use models to acquire more profound insights into challenging cases. Apply to perform tasks better | Dashboards that provide visualizations showing which actions can issue more effectively |
| Unions and regulatory bodies (e.g., WorkSafe) | | Various techniques can be applied to show how the model works (See Figure 5) |
| Clients and financiers | Identify actionable recourse to improve model decisions for future projects and/or asset management | Feature attribution and/or reason codes show which variables significantly impacted project success or failure. If used for asset management, how will the system be improved? |

Adapted from Surkov *et al*. (2022)

Regarding the process, two key issues come to the fore: (1) the development, implementation, and operation and maintenance of an AI model (e.g., DevOps[4] in conjunction with AIOps[5]) with explainability designed into a new application's functioning (Figure 4). In the case of existing AI models, explainability may be retrospectively added, though whether stakeholders would have confidence, trust, and understanding of the explanations developed remains questionable as they would have had no input into the system's design; and (2) the ensuing change process that typically manifests with the introduction of technology and new ways of working. The outcomes are the realization of benefits, which are strategic, operational, and tactical.

An essential feature of our framework is the organization-project dyad, two distinct yet interdependent contexts. Even though the construction organization will develop an AI system for their benefit, they will need to accommodate their stakeholders, particularly those from the project, to produce and evaluate an XAI model. We suggest that a series of interventions during the evaluation process are performed, including participatory workshops, survey feedback, and information events (Fridlich *et al*., 2015). Such intervention elements should be incorporated into the XAI systems architecture and development. Discussing and understanding the rationale behind an algorithm's purpose and selecting and assessing explanations are critical areas for consideration during an AI model's design and development (Figure 3).

Currently, propriety and bespoke AI models being deployed in construction have been typically developed and evaluated in isolation of stakeholder input. For example, to the authors' knowledge and based on our experience in developing bespoke safety systems using DL and computer vision, we have not observed the input of stakeholders such as unions and subcontractors in their design, use, and evaluation (e.g., Ding *et al*., 2018; Fan *et al*., 2019; Wei *et al*., 2019; Fang *et al*., 2020a; Fang *et al*., 2021; Fang *et al*., 2022). Issues such as privacy and security, to name a few, remain unaddressed issues in the construction literature with regard to applying AI models during construction (Fang *et al*., 2020b; Liu *et al*., 2022). If strides are to be made for the widespread acceptance of AI models for monitoring safety performance in construction, XAI and the inclusion of stakeholder desiderata will be required.

---

[4] DevOps (i.e., 'development" and "operations') is the combination of practices and tools designed to increase an organization's ability to deliver software applications and services faster than traditional development processes. This speed enables organizations to better serve their customers and compete more effectively (Bass *et al*., 2015).

[5] AIOps is defined as "platforms utilize big data, modern Machine Learning and other advanced analytics technologies to directly and indirectly enhance IT operations (monitoring, automation and service desk) functions with proactive, personal and dynamic insight. AIOps platforms enable the concurrent use of multiple data sources, data collection methods, analytical (real-time and deep) technologies, and presentation technologies" (Lerner, 2017).

However, if AI models are to be explainable, then the intervention elements must be designed and planned so that understandable explanations can be established throughout their development process by establishing and enacting a governance model (Figure 4) that considers accountability and responsibility (responsibility explanation). Two philosophies for governance exist (Arrieta *et al*., 2020): (1) where committees review and approve AI developments and amendments; and (2) the self-responsibility of employees. In the case of committees, a construction organization must resolve issues with whom to involve in the intervention, when, how, and who will lead and manage the process (Figure 2). Indeed, this will be a dilemma as not all stakeholders are equal. The construction organization will naturally want control over its investment and the benefits to be realized.

Faced with this dilemma, we suggest establishing a steering group of stakeholders led by a 'technological maestro' employed by the construction organization to manage the intervention process. Such a maestro would aim to be impartial, possess in-depth AI knowledge, and equally understand construction. Additionally, they would have a questioning mindset and know what to ask in order to ensure an AI model is explainable, interpretable, and transparent (Westrum, 2004). Adopting a series of interventions with stakeholders to develop XAI is outside the remit of construction organizations as it would be seen as a distraction from their core business. However, as construction organizations rely more on real-time information for decision-making, such interventions will become an inevitable part of the XAI evaluation process during its implementation, especially to assess a model's performance. Moreover, an AI model's users should be empowered and encouraged to openly question and seek explanations about its outputs.

During the operation and maintenance of an AI model, monitoring its efficacy (e.g., benchmarking) and the changes that have been made to work practices is necessary. Moreover, the AI model's performance and ability to provide understandable explanations and benefits will need to be assessed and managed. Ensuring an XAI system can achieve a sustained and long-term impact will require a governance model to be enacted to ensure 'Responsible AI'[6] principles (e.g., explainability, fairness, human agency, privacy, security, and safety) are understood, adopted, and managed. The upshot is the need to strike a balance between (Arrieta *et al*., 2020): (1) cultural and organizational changes to enforce such principles across all processes endowed with AI functionalities; and (2) the feasibility and compliance of the implementation of such principles with the information technology, assets, policies, and resources available within the construction organization. By raising awareness around the principles and values of 'Responsible AI,' construction organizations can justify and position the need for embracing XAI in the workplace and their projects.

---

[6] The Department of Industry Science and Resources in Australia, for example, has established AI ethics principles, which can be found at: https://www.industry.gov.au/publications/australias-artificial-intelligence-ethics-framework/australias-ai-ethics-principles. Similarly, the OECD has established AI Principles which are available at: https://oecd.ai/en/ai-principles

## 5.0 Implications for Research

Our proposed XAI evaluation framework is conceptual. It provides the first attempt to understand how construction organizations can acquire the benefits of implementing AI models and ensure their successful adoption in business functions and projects. The concept of XAI is evolving rapidly. Instead of waiting for it to mature, construction organizations are well placed to take advantage of its benefits as they currently have little, if any, reliance on AI to perform their operations. Being an early adopter of XAI would not only provide a construction organization with a competitive advantage, but it could also influence its development, provide early access to solutions for common problems, and help the industry with its performance and productivity problems.

To assume that a construction organization would engage in altruistic behavior and take on the challenge of becoming an 'XAI thought leader' is optimistic. After all, they are still trying to understand *what*, *why,* and *how* the benefits of AI can be realized. However, we are hopeful that some construction organizations will take on this challenge as our *weltanschauung* is oriented towards a glass 'half full' rather than 'half empty'. There is a need for research to focus on several areas to facilitate and support construction organizations that choose to embrace XAI and evaluate its implementation.

### *5.1     Utilize Post-hoc Type Methods*

A cursory review of the XAI literature reveals that there is no unified concept of explainability, though there is some consensus on its basic precepts (Arrieta *et al*., 2020; Amarasinghe *et al*., 2021; Angelov *et al*., 2021; Belle and Papantonis, 2021; Love *et al*., 2022). To drive the field of XAI forward, there is a need to have a common understanding of explainability. Without it, a standard structure for XAI systems cannot be developed, and new techniques and methods cannot be established.

Put simply, explainability is defined as the ability of an AI model to make its functioning clearer and more understandable to the audience it serves. However, ML and DL studies in construction have tended to overlook the need to explain why and how their models provide the outcome they recommend. While the field of XAI seeks to build a shared understanding of explainability, AI research in construction, in the meantime, can draw on existing tools and techniques (e.g., post-hoc type methods) to demonstrate a model's utility and help users understand its functioning and outputs (Figure 5). As a result, this will enable AI models to be evaluated to determine whether it does what it is supposed to do. We suggest that our proposed XAI framework can be used to facilitate the evaluation process. However, it may need to be expanded to accommodate the requirements and needs of the organization.

### 5.2 *Measures of Performance: Enabling 'Management by Fact'*

Metrics help add clarity to the evaluation process providing an ability to manage the effective implementation of XAI by fact. In our evaluation framework, metrics would allow for a meaningful comparison of how well an AI model fits an explanation, but they have yet to be established in the literature (Arrieta *et al*., 2020; Langer *et al*., 2021; Khosravi *et al*., 2022). Without such metrics, our proposed XAI evaluation framework cannot be wholly operationalized.

Needless to say, metrics do exist for evaluating the performance of ML and DL models, which include: (1) classification metrics (e.g., accuracy, precision, recall, and F-1 score); (2) regression metrics (i.e., mean squared root and root mean squared root and mean absolute error); and (3) computer vision (i.e., peak to signal to noise ratio, structural similarity index, and Intersection over Union). However, XAI metrics need to go beyond simply measuring a model's performance to evaluating its effectiveness, usefulness, and satisfaction with explanations. They also should measure "the improvement of the mental model of the audience induced by model explanations, the impact of explanations on the model's performance, and the trust and reliance of the audience" (Arrieta *et al*., 2020: p. 100).

Even though measures of explanation have been propagated in the literature "(e.g., goodness checklist, explanation satisfaction scale, elicitation method for mental models, computational measures for explainer fidelity, explanation of trustworthiness and model reliability)" they are subjective (Arrieta *et al*., 2020: p. 100). Such measures can be applied within our proposed framework. They should be trailed in future research, but the fact remains that quantifiable metrics are still required. Once we can comprehend and determine how qualitative measures can be used to provide explanations, then strides can be made to develop quantifiable metrics.

**6.0    Conclusion**

Research examining XAI in construction is in its infancy. A considerable amount of research is required to prepare construction organizations for its adoption and effective implementation in the future. Integral to the adoption of XAI is having a formalized evaluation process to ensure it is successfully deployed and its benefits are realized. Nevertheless, within the extant literature, no XAI evaluation frameworks have been developed as there is a lack of clarity around the concept of explainability. However, there is agreement about the basic precepts of XAI (i.e., explainability, interpretability, and transparency).

We believe construction organizations are well-positioned to take advantage of XAI as AI has yet to establish itself as an everyday technology on-site. Thus, construction organizations are presented with a clean sheet which they can work from to develop explainable AI models and harness their benefits. To help construction organizations understand how they can take advantage of XAI, we have developed an evaluation framework guided by questions addressing the following: (1) Why is the evaluation being done? (2) What is being evaluated? (3) Who affects the evaluation? (4) When is the evaluation taking place? (5) How is the evaluation to be carried out? and (6) what are its benefits? Our evaluation framework considers XAI a meta-human system with stakeholders critical to its successful deployment.

We need to point out that our evaluation framework is conceptual. While explainability has not received attention in AI studies in construction, we suggest future research consider utilizing current post-hoc methods to begin the effort toward developing an understanding and explanation of the functioning and outputs of ML and DL models. We also suggest that research should evaluate explanations using existing qualitative approaches and then progress toward developing quantifiable metrics. To this end, we finally note that our framework offers researchers a frame of reference to examine the success or failure of XAI in construction.

**Acknowledgments**

We want to acknowledge the financial support of the *Australian Research Council* (DP210101281) and (DE230100123). Also, we would like to acknowledge the financial support of the Alexander von Humboldt-Stiftung, and the National Natural Science Foundation of China (Grant No. U21A20151).